# A Hybrid Approach to Reasoning with Partially Elicited Preference Models

**Vu Ha**★
★ Decision Systems and Artificial Intelligence Lab
Dept. of EE&CS
University of Wisconsin-Milwaukee
Milwaukee, WI 53211
{vu,haddawy}@cs.uwm.edu

**Peter Haddawy**★†
† Intelligent System Lab
Faculty of Science & Technology
Assumption University
Bangkok 10240, Thailand
haddawy@isl.s-t.au.ac.th

## Abstract

Classical Decision Theory provides a normative framework for representing and reasoning about complex preferences. Straightforward application of this theory to automate decision making is difficult due to high elicitation cost. In response to this problem, researchers have recently developed a number of qualitative, logic-oriented approaches for representing and reasoning about preferences. While effectively addressing some expressiveness issues, these logics have not proven powerful enough for building practical automated decision making systems. In this paper we present a hybrid approach to preference elicitation and decision making that is grounded in classical multi-attribute utility theory, but can make effective use of the expressive power of qualitative approaches. Specifically, assuming a partially specified multilinear utility function, we show how comparative statements about classes of decision alternatives can be used to further constrain the utility function and thus identify sup-optimal alternatives. This work demonstrates that quantitative and qualitative approaches can be synergistically integrated to provide effective and flexible decision support.

## 1 INTRODUCTION

Within the field of automated decision making, similar to the early days when probability theory was considered epistemologically inadequate, utility theory these days faces several epistemological problems of its own. In particular, it is often quite difficult to elicit the required utility function, especially when the outcomes of the decisions are complex. While techniques exist for eliciting a complete utility function from a user, doing so may be neither practical nor desirable. First, a large elicitation overhead may not be commensurate with the task at hand. Second, since people's preferences tend to change over time, we may wish to represent only that core of preferences that is relatively stable over some desired time period. Thus we would like to develop techniques for partially eliciting preferences and for reasoning with partially specified preferences in order to eliminate suboptimal decision alternatives.

Practitioners of decision theory have addressed the issue of eliciting utility functions by developing a comprehensive framework, generally known as multi-attribute utility theory (MAUT) [10]. Lying at the heart of MAUT is the notion of *utility independence*, one of the first notions introduced to exploit qualitative, structural aspects of preference. Suppose that decision outcomes can be described by a set $X = \{X_1, X_2, ..., X_n\}$ of attributes, meaning that an outcome $x$ is a value assignment $(X_1 = x_1, X_2 = x_2, ..., X_n = x_n)$ to the attributes, a set $Y \subset X$ is said to be *utility independent* of its complement $X - Y$, or *UI* for short, if the preference over probability distributions $P$ whose marginals over $X - Y$ are a fixed, degenerate distribution $P_{X-Y}$ does not depend on $P_{X-Y}$.

Utility independence occurs quite often in real-life decision making situations, and in general can be detected easily. When a set of attributes $Y$ is UI, a simple theorem shows that we can write the utility function $u(x)$ as an expression that consists of two functions over $X - Y$ and one function over $Y$, achieving a reduction of dimensionality (and hence complexity). In particular, if every subset $Y$ of $X$ is UI, a condition called *mutual utility independence (MUI)*, then we can write $u(x)$ either in a multiplicative form:

$$u(x) = \prod_{i=1}^{n}(1 + k_i u_i(x_i)),$$

or in an additive form:

$$u(x) = \sum_{i=1}^{n} k_i u_i(x_i),$$

where $u_i$ are so-called sub-utility functions that capture the decision maker's preference with regard to attribute $X_i$ when holding the attributes $X_j, j \neq i$ at some fixed level, and $k_i$ are constants that ensure proper global scaling [1].

Whenever MUI is applicable, the utility function can be obtained in the following two steps:

1. *Determining the individual objectives.* The sub-utility functions $u_i, i = 1, ..., n$ are assessed. For this purpose, the decision maker is asked to rank decision consequences that have the values for $X_j, j \neq i$ fixed. This step is relative simple. For example, when $X_i$ is propositional, the cost of assessing $u_i$ is zero; $u_i$ is either 0 or 1, depending on $x_i$ being the inferior or superior value.

2. *Determining the tradeoffs.* The scaling coefficients $k_i, i = 1, ..., n$ are assessed, typically by determining the relative ratios for certain pairs of the scaling coefficients. For this purpose, the decision maker is asked to rank decision consequences that differ in the two corresponding attributes.

In real-world situations, however, MUI is often not applicable. For example, we may only be able to structure the outcome space in terms of attributes that are individually UI, i.e., $\{X_i\}$ is UI, $i = 1, 2, ..., n$. In such cases, the utility function takes on the *multi-linear* (MLUF) form:

$$u(x) = \sum_{\emptyset \neq Y \subseteq X} k_Y \cdot \prod_{X_i \in Y} u_i(x_i), \qquad (1)$$

where $u_i$ are sub-utility functions, and $k_Y, \emptyset \neq Y \subseteq X$ are scaling coefficients.

In the case when the utility function is assumed to be multi-linear, while assessing the sub-utility functions $u_i$ is still relatively easy, assessing a total of $2^n - 1$ scaling coefficients is quite daunting [2]. This complexity poses a difficult dilemma to the decision analyst: she can either work with an additive or a multiplicative function even when evidence suggests that MUI is violated, in effect obtaining an approximate model of the decision maker's preference, or work with a MLUF, placing a sizable elicitation burden on herself and on the decision maker.

Our perception is that in real-world applications of decision theory, decision analysts often choose the former option: to assume MUI. In this paper, we propose to study MLUFs [3]. Because complete elicitation of MLUFs is impractical, we set out to develop techniques to reason with *partially elicited* MLUFs. In particular, these techniques are designed to identify sub-optimal decision alternatives without assessing all the $2^n - 1$ scaling coefficients (expected utilities thus are not explicitly computed). The premises of these techniques are the following:

(i) The set of available decision alternatives is finite. Each decision alternative results in a completely specified probability distribution over the states of the world.

(ii) The decision maker's utility function is multi-linear. The sub-utility functions have already been elicited, but the scaling coefficients are still unknown.

(iii) The decision maker can provide a set of preferential comparisons of the form $p_j \preceq q_j, j$, where $p_j$ and $q_j$ are decision consequences [4].

In Section 2, we show that the above assumptions can be captured by a polyhedral cone that constrains the unknown scaling coefficients $k_Y$. We then use standard optimization techniques to deduce further preferential information such as *induced dominance* and *potential optimality* from this constraint. An inherent benefit of these techniques is that they can be used to detect inconsistency in the prefence information elicited from the user. (We note that previously, Hazen presented a preference cone approach for reasoning with partially specified additive or multiplicative utility functions [9], which is similar to the approach presented in this paper.)

The key assumption in this approach is (iii), which assumes that the decision maker can provide preferential comparisons between (real or fictious) decision consequences. This assumption is an integral part of interactive approaches to decision making [12, 9, 11].

---

[1] Usually, the constants $k_i$ are scaled so that both $u$ and sub-utility functions $u_i$ have the range [0, 1].

[2] In fact, in the case when there are more than 3 attributes, assessment of MLUFs is usually abandoned [10].

[3] Bacchus and Grove also take this stand: "We conjecture that [multilinear utility models] might be worth studying in the context of artificial intelligence applications, and in particular for giving a better decision-theoretic account of goal" [2].

[4] In this paper, we use the term "decision consequences" to indicate both *outcomes*, which are consequences of decisions with certainty, and *prospects*, which are consequences of decisions with uncertainty. Prospects are probability distributions over outcomes.

The rationale for making this assumption is that there are certain circumstances where it might be easier for the decision maker to express preferences among decision consequences than to introspect about the attributes describing each one. For example, in expressing preferences about movies, most people can readily express their preferences over two films they have seen in the past but may have difficulty describing preferences over attributes like director, leading actor, or costume designer. In fact, most people would not even recognize the names of the costume designers, even when they may have a preference for films with nice costumes.

The more comparisons the decision maker can provide, the more conclusive inference can be made. In particular, if the decision maker can provide a succint, qualitative statement about her preference that implicitly encoded a *set* of comparision statements, then we may be able to quickly identify a large set of sub-optimal alternatives.

To capture such preferential statements and derive efficient inference mechanisms using them is one of the aims of the field of qualitative decision theory. Recent work from this field has attempted to address the elicitation problem by providing formal languages in which partial preference information can be conveniently expressed [5, 6, 15, 3, 1]. For example, the languages proposed by Doyle and Wellman [6] and by Tan and Pearl [15] attempt to provide a semantic to *ceteris paribus* (all else being equal) comparative statements. These are preferential statements concerning classes of decision consequences.

While these languages have successfully addressed a number of expressiveness issues, the inferential mechanisms available have not been sufficiently powerful for building practical decision making systems. In Section 3, we propose using ceteris paribus comparative statements, as presented in [6], as a means to represent comparative statements made by the decision maker, to be used in conjunction with the assumptions (i), (ii), and (iii) described above. This combination provides us with a flexible representation for eliciting preferences and an inferrential mechanism to effectively eliminate decision alternatives. The resulting hybrid framework is intended to strike a balance between logic-oriented (generally too weak), and numeric-oriented (generally too cost-intensive) approaches.

The rest of this paper is organized as follows. In Section 2, we develop the preference cone framework to reason with partial MLUF and pairwise comparisons. In Section 3, we propose to intergrate qualitative preferential comparisons to the this framework and provide an example to illustrate this idea. We finish with discussion of related work and future research issues.

## 2 REASONING WITH PARTIALLY ELICITED MULTI-LINEAR UTILITY FUNCTIONS USING POLYHEDRAL CONES

In this section we explore the idea of using explicit pairwise comparisons of decision consequences to identify sub-optimal alternatives, as outlined in the Introduction. The premises of this analysis are assumptions (i), (ii), (iii).

### 2.1 LINEAR CONSTRAINTS ON THE SCALING COEFFICIENTS OF MLUFS

First, note that the multilinear form of the utility function, as formalized in Equation (1) does not fully capture the assumption that the attributes $X_i$ are UI; the multilinear form is only a necessary but not sufficient condition for $X_i$ to be UI. We need to add constraints on the scaling coefficients $k_Y$ in order to obtain a necessary and sufficient condition.

Take, for example, the assumption that $X_i$ is UI. Let $t_Y(x) = \prod_{X_j \in Y} u_j(x_j)$, $\emptyset \neq Y \subseteq X$. The multi-linear utility function $u(x)$ in Equation 1 can be written as

$$\begin{aligned}u(x) &= \sum_{\emptyset \neq Y \subseteq X} k_Y t_Y(x) \\ &= \left( \sum_{Z \subseteq X - \{X_i\}} k_{\{X_i\} \cup Z} t_Z(x) \right) u_i(x_i) \\ &\quad + \sum_{\emptyset \neq Y \subseteq X - \{X_i\}} k_Y t_Y(x),\end{aligned}$$

and thus can be viewed as a linear function of $u_i(x_i)$. Thus to say that $X_i$ is UI is equivalent to say that the coefficient for $u_i(x_i)$ in this linear function must be non-negative. Formally, this means

$$\sum_{Z \subseteq X - \{X_i\}} k_{\{X_i\} \cup Z} t_Z(x) \geq 0. \quad (2)$$

Moreover, this inequality must be satisfied for *any* value assignment to the attributes in the set $X - \{X_i\}$. Inversely, if this constraint is satisfied, then $X_i$ is UI. In other words, the utility independence of the attributes $X_i$ is precisely captured by the multilinear form as in Equation ( 1), with the additional linear, homogeneous constraints about the scaling constants, as expressed in Inequality ( 2).

To further simplify the expositions, we introduce the following notations. Let **k** denote the $d$-dimensional ($d = 2^n - 1$) vector with components $k_Y, \emptyset \neq Y \subseteq X$. For any $i = 1, 2, ..., n$, let $\mathbf{s}^i$ denote the same-dimension vector whose components are functions $s^i_Y : X \to \Re, \emptyset \neq Y \subseteq X$, defined as:

$$s^i_Y(x) = \begin{cases} 0 & \text{if } X_i \notin Y \\ -t_{Y-\{X_i\}}(x) & \text{otherwise.} \end{cases}$$

Note that the functions $s^i_Y(x)$ do not depend on $x_i$, the $i$-th component of $x$. Inequality (2) can thus be written as

$$\langle \mathbf{k}, \mathbf{s}^i(x) \rangle \leq 0, \forall i = 1, 2, ..., n; x,$$

where $\langle ., . \rangle$ denotes the inner product of two vectors.

One may at first think that these constraints are by themselves strong enough to imply non-trivial preferences over decision alternatives. However, the results that we were able to obtain in our previous work [7], presented in the theorem below, suggest that further preferences can be deduced in only very special cases.

**Theorem 1** *Let $p$ and $q$ be two decision alternatives. Further assume that $E_p[u_i(x_i)] \leq E_q[u_i(x_i)], \forall i$, i.e., $q$ would be preferred to $p$ if we took the sub-utility function $u_i(x_i)$ as our overall utility function. From these so-called* local dominance *conditions, we can infer overall dominance, i.e. $p \preceq q$ if either:*

*(a) the utility function is additive, or*

*(b) in the probability distributions $p$ and $q$, the attributes $X_i$, when viewed as random variables, are probabilistically independent.*

*Furthermore, the inference is not sound if the utility function is multiplicative.*

## 2.2 LINEAR CONSTRAINTS FOR PAIRWISE COMPARISONS

In order to be able to infer a pairwise preference, we need to impose very strong conditions, either about the form of the utility function (it must be additive), or about alternatives (they must be probabilistically independent), in addition to having the local dominances. When these conditions do not hold, we need other sources of preferential information in order to be able to identify sub-optimal alternatives and to narrow down the set of candidate alternatives. One such source is pairwise comparison statements made by the decision maker.

Note that the statement $p \preceq q$ translates into the following inequalities:

$$E_p[u(x)] \leq E_q[u(x)] \Leftrightarrow$$
$$E_p\left[\sum_Y k_Y t_Y(x)\right] \leq E_q\left[\sum_Y k_Y t_Y(x)\right] \Leftrightarrow$$
$$\sum_Y k_Y E_p[t_Y(x)] \leq \sum_Y k_Y E_q[t_Y(x)].$$

Now, denoting $t_Y(p) = E_p[t_Y(x)] = \sum_x p(x) t_Y(x)$, and $t_Y(q) = E_q[t_Y(x)] = \sum_x q(x) t_Y(x)$, we then have

$$p \preceq q \Leftrightarrow \sum_Y k_Y t_Y(p) \leq \sum_Y k_Y t_Y(q)$$
$$\Leftrightarrow \langle \mathbf{k}, \mathbf{t}p - \mathbf{t}q \rangle \leq 0,$$

where $\mathbf{t}p$ (respectively $\mathbf{t}q$) denotes the $(2^n - 1)$-dimensional vector whose components are $t_Y(p)$ (respectively $t_y(q)$). This last inequality is also a linear, homogeneous constraint over the unknown constants $k_Y$.

## 2.3 SOME BASIC CONCEPTS OF CONVEX CONE ANALYSIS

Before we continue our analysis, we provide a brief review of some basic concepts of convex cone analysis.

**Polyhedra, Cones, and Polyhedral Cones**

The $d$-*dimension Euclidean Space* is the vector space $\Re^d$ equipped with the inner product $\langle \rangle$. Given a vector $\mathbf{n}$, and $\alpha \in \Re$, the set $\mathbf{n}_\alpha = \{\mathbf{x} | \langle \mathbf{n}, \mathbf{x} \rangle = \alpha\}$ is called a *hyperplane*, the set $\mathbf{n}^-_\alpha = \{\mathbf{x} | \langle \mathbf{n}, \mathbf{x} \rangle \leq \alpha\}$ is called a *closed halfspace* with *outward normal* $\mathbf{n}$. For simplicity of notations, the subscript $\alpha$ is omitted when $\alpha = 0$. The intersection of a finite number of closed halfspaces is called a *polyhedron*.

A set $W \subseteq \Re^d$ is called a *cone with apex* 0 if $\lambda \mathbf{x} \in W$ whenever $\lambda \geq 0$ and $\mathbf{x} \in W$. A set $W$ is a cone with apex $\mathbf{a} \in \Re^d$ if $W - \mathbf{a} := \{\mathbf{x} - \mathbf{a} | \mathbf{x} \in W\}$ is a cone with apex 0. In this paper, cones are all 0-apexed, unless indicated otherwise. Given a set $W \subseteq \Re^d$, the set of all points that can be expressed as non-negative linear combinations of points of $W$ can be shown to be a convex cone, and is denoted by $C_W$. This cone is called the convex cone generated by $W$ and can be equivalently defined as the smallest convex cone containing $W$. A cone that is also a polyhedron is called a *polyhedral cone*. It is well-known that polyhedral cones are precisely convex cones generated by finite sets of points, and can be shown to be *closed*.

**Dual Cones**

Given a set $W \subseteq \Re^d$, let $W^*$ be defined as $W^* := \{y|\langle x,y\rangle \leq 0, \forall x \in W\}$. $W^*$ is easily shown to be a convex cone and is refered to as the *dual cone of $W$*. For example, if $W$ contains a single point $\mathbf{n}$, then the dual cone of $W$ is $\mathbf{n}^-$, the closed halfspace with outward normal $\mathbf{n}$.

The following theorem is standard in convex cone analysis.

**Theorem 2** *Let $W \subseteq \Re^d$.*

*(a) $W^* = (C_W)^*$. Any set and the convex cone it generates share the same dual cone.*

*(b) $W^{**} := (W^*)^* = \overline{C}_W$. The dual cone of the dual cone of $W$ is equal to the closure of the convex cone it generates. In particular, if $W$ is finite, then $W^{**} = C_W$.*

## 2.4 INDUCED DOMINANCE, POTENTIAL OPTIMALITY, AND INCONSISTENCY DETECTION

Recall that the analysis in Subsections 2.1 and 2.2 shows that the assumptions (i), (ii), and (iii) described in the Introduction can be precisely captured by the following inequalities

$$\begin{cases} \langle \mathbf{k}, \mathbf{s}^i(x) \rangle & \leq 0, \forall i, x \\ \langle \mathbf{k}, tp_j - tq_j \rangle & \leq 0, \forall j. \end{cases}$$

Assuming that the domain of each attribute $X_i$ is finite, the above inequalities are equivalent to a finite set of linear homogeneous constraints over the scaling coefficients $k_Y$. From now on, we denote these constraints as follows $\{\langle \mathbf{k}, \mathbf{w}_j \rangle \leq 0 | j = 1..m\}$, and denote $W = \{\mathbf{w}_j | j = 1..m\}$. Thus, the utility function $u$, when represented as a vector $\mathbf{k}$ with coordinates $k_Y$, must lie in the intersection of the closed halfspaces with outward normal $\mathbf{w}_j$:

$$\mathbf{k} \in K := \bigcap_{j=1}^m \mathbf{w}_j^-.$$

This intersection is a polyhedral cone, which is the *dual cone $K = W^*$ of $W$*. Using Theorem 2, we have that $K^* = W^{**} = C_W$, i.e., the dual cone of $K$, the set of admissible $\mathbf{k}$, is the polyhedral cone generated by the constraint vectors $\mathbf{w}_j$ (see Figure 1).

**Induced Dominance**

The above analysis immediately leads to the following result that can be used to test for induced dominance.

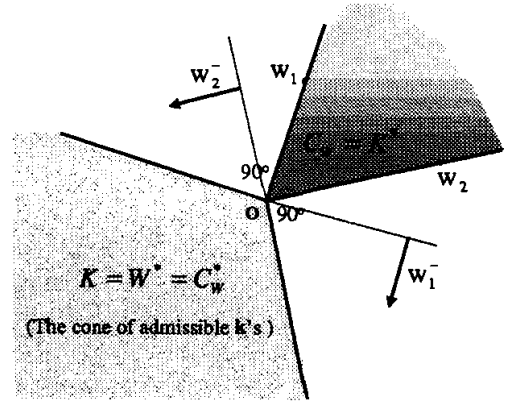

Figure 1: A 2-dimensional illustration of Theorem 3. Here $W = \{\mathbf{w}_1, \mathbf{w}_2\}$.

**Theorem 3** *Given a pair of alternatives $(p,q)$, we can deduce $p \preceq q$ iff $tp - tq \in C_W$. The "only if" part means that if $tp - tq \notin C_W$, then there exists $\mathbf{k} \in K$ such that $\langle \mathbf{k}, tp - tq \rangle > 0$, i.e. $p \succ q$ under such $\mathbf{k}$.*

Below is an algorithm for testing induced dominance. This "inference rule" is sound and complete based on the above theorem.

**Algorithm 1** .
**Input:** *Two prospects $p$ and $q$.*
**Output:** *Return*

$$\begin{cases} 1 & \text{if } p \preceq q, \\ -1 & \text{if } q \preceq p, \\ 0 & \text{if the relationship between} \\ & p \text{ and } q \text{ cannot be determined}. \end{cases}$$

*1. Determine a set of generators $\{\mathbf{k}_l|l\}$ of the polyhedral cone $K$.*

*2. Return 1, if $\langle \mathbf{k}_l, tp - tq \rangle \leq 0, \forall l$.*

*3. Return $-1$, if $\langle \mathbf{k}_l, tq - tp \rangle \leq 0, \forall l$.*

*4. Return 0, otherwise.*

**Complexity Analysis**

The complexity of Algorithm 1 is determined by the complexity of determining the generators $\{\mathbf{k}_l|l\}$ of the polyhedral cone $K$ (step 1). These vectors are outward normal vectors of the $(d-1)$-dimension facets of the polyhedron $C_W$. This problem is essentially the same problem as finding the facet normals of the convex hull of a set of points in the $(d-1)$-dimension space, which can be solved in time $O(m^{\lceil d/2 \rceil - 1})$ using algorithms from computational geometry [14]. Here, $m$ is the number of the elements of $W$, and $d = 2^n - 1$ is the dimension of the space (which is the number of the scaling coefficients).

**Potential Optimality**

We can also test for potential optimality in a straightforward way. Given a set of decicision alternatives $\{a_1, a_2, ..., a_m\}$ resulting in the corresponding probability distributions $\{p_1, p_2, ..., p_m\}$, decision alternative $a_r$ is potentially optimal if $tp_r - tp_s \notin C_W, \forall s = 1, 2, ..., m, s \neq r$, or, equivalently, the polyhedral cone generated by $\{tp_r - tp_s | s = 1, 2, ..., m, s \neq r\}$ intersects with $C_W$ at the origin only.

To conclude this section, we note that in complex decision making problems, the decision maker may easily exhibit inconsistent preferences. For example, she may assert a set of comparative statements that results in an empty preference cone ($K = \emptyset$). In such situations, we would like the system to restore consistency by eliminating one or more "problematic" comparative statements. But how to identify such problematic statements is an open question and needs further research and experiment.

## 3  INTERGRATING QUALITATIVE COMPARATIVE STATEMENTS

In the previous section we have shown how comparative statements about decision consequences can be exploited to test for induced dominance and potential optimality. Since there are $2^n - 1$ unknown scaling coefficients, chances are that we would need an inordinately large number of pair-wise comparisions from the decision maker in order to make useful inferences.

Qualitative preference logics such as those proposed by Doyle and Wellman [5, 6] and Tan and Pearl [15] provide languages that can express comparative statements about classes of decision consequences. Such a qualitative expression of preferences gives us a large number of pair-wise preferences among individual decision conequences, which can be used to effectively constrain the space of utility functions.

In this section we give an example to illustrate this idea. In this example, the state space has three propositional attributes $\{X_1, X_2, X_3\}$ with domains $\{0, 1\}$. Suppose also that the attributes are utility independent in such a way that for each attribute, 1 is the prefered value. This means that the sub-utility functions $u_i(x_i)$ for the attributes are given by $u_i(x_i) = x_i$. The overall utility function can be written as

$$u(x_1, x_2, x_3) = k_1 x_1 + k_2 x_2 + k_3 x_3 + k_{12} x_1 x_2 + k_{13} x_1 x_3 + k_{23} x_2 x_3 + k_{123} x_1 x_2 x_3.$$

Next, we translate the assumption about utility independence of the attributes into constraints about the scaling constants. Let us consider attribute $X_1$. To say that $X_1$ is UI is equivalent to say that the overall utility function

$$u(x_1, x_2, x_3) = (k_1 + k_{12} x_2 + k_{13} x_3 + k_{123} x_2 x_3) x_1 + k_2 x_2 + k_3 x_3 + k_{23} x_2 x_3,$$

when viewed as a linear function of $x_1$ must have positive coefficient for $x_1$, i.e.

$$k_1 + k_{12} x_2 + k_{13} x_3 + k_{123} x_2 x_3 > 0, \forall x_2, x_3.$$

Considering all value assigments for $x_2$ and $x_3$, the UI assumption for $X_1$ implies the following inequalities involving the scaling constants:

$$\begin{cases} k_1 & > 0 \\ k_1 + k_{12} & > 0 \\ k_1 + k_{13} & > 0 \\ k_1 + k_{12} + k_{13} + k_{123} & > 0 \end{cases} \quad (3)$$

The utility independence for $X_2$ and $X_3$ can be expressed similarly:

$$\begin{cases} k_2 & > 0 \\ k_2 + k_{12} & > 0 \\ k_2 + k_{23} & > 0 \\ k_2 + k_{12} + k_{23} + k_{123} & > 0 \end{cases} \quad (4)$$

$$\begin{cases} k_3 & > 0 \\ k_3 + k_{13} & > 0 \\ k_3 + k_{23} & > 0 \\ k_3 + k_{13} + k_{23} + k_{123} & > 0 \end{cases} \quad (5)$$

Note that the inequality $k_1 > 0$ is equivalent to the comparison statement $(0, 0, 0) \prec (1, 0, 0)$, and the inequality $k_1 + k_{12} > 0$ is equivalent to $(0, 1, 0) \prec (1, 1, 0)$.

Now suppose that the decision maker states that she prefers to have $(x_1 = 1, x_2 = 0)$ to $(x_1 = 0, x_2 = 1)$, all else being equal, i.e. regardless of the value of $x_3$. This statement is equivalent to the following constraints:

$$\begin{cases} k_1 & > k_2 \\ k_1 + k_{13} & > k_2 + k_{23} \end{cases} \quad (6)$$

Using Algorithm 1 with the Constraints (3), (4), (5), and (6), we will be able to obtain that, for example,

$$(1,0,1) \succ \{(.5,(0,1,1)),(.3,(1,0,0)),(.2,(0,1,0))\},$$

where the right hand side is a probability distribution giving the probabilities .5, .3, .2 to the states $(0,1,1),(1,0,0),(0,1,0)$, respectively.

## 4 RELATED WORK

The idea of representing partial preference information using polyhedral cones has appeared in work in the field of Multiple Criteria Decision Making (MCDM). In this work, decision alternatives are scored according to a finite number of criteria, and the overall score for each alternative is a (value) function of the individual scores. In this sense, all decision alternatives result in *certain* outcomes that have scores as attributes. In contrast, in this paper, the consequences of decisions are uncertain. Furthermore, in work in MCDM, the value function is usually assumed to have some tractable form such as (in increasing order of generality) linear [16], quasiconcave [12, 13], or monotonic [11]. In this paper, the decision maker's utilty function is assumed to have multi-linear form. Since multi-linear functions can be non-monotonic (see Theorem 1), it is not immediately clear if the preference cone techniques from the work in MCDM mentioned above can be used in our approach.

## 5 DISCUSSION

Classical Decision Theory provides a normative framework for representing and reasoning about complex preferences. Straightforward application of this essentially quantitative theory to automate decision making is difficult due to high cost of eliciting utility functions. Recent work from the field of qualitative decision theory offers several alternative solutions. These approaches focus on developing formal languages that can express qualitative, partial preference information. However, the inference mechanisms offered by these languages remain rather weak.

It is thus highly desirable to develop a framework that can exploit different strengths of these different quantitative and qualitative approaches. In this paper, we attempt to provide such a framework. Assuming a multi-linear utility function with known sub-utility functions, we show how ceteris paribus comparative statements by the decision maker can be used to infer further preferential information such as induced pairwise preference and sub-optimality.

There are several issues that we plan to address in this framework. The first issue is efficiency. Note that the time complexity of Algorithm 1 is exponential ($O(m^{\lceil d/2 \rceil -1})$). The base of this upper bound, which is the number of the constraint vectors $\mathbf{w}_j$, can be large. A possible solution for this is to investigate ways to *effectively* use sets of comparisions, represented by qualitative logical constructs. For example, a ceteris paribus comparison, which is equivalent to a *set* of comparisons between individual decision consequences, can sometimes be captured by a *single* linear inequality involving certain (but not all) scaling coefficients. This would result in fewer constraint vectors. With regard to the exponent of the upper bound, since in practice, the number of attributes ($n$) is usually very small (3-8), we expect the exponent ($\lceil d/2 \rceil - 1$) not to exceed 127. We can also try to exploit further utility independencies, if applicable, to reduce the number of scaling coefficients and to reduce the exponent. For example, if $Y$ is UI of $X - Y$ and $|Y| = r$, then the number of scaling coefficients can be reduced to $2^r + 2^{n-r+1} - 4$. If both $Y$ and $X - Y$ are UI of the other, then the number can be reduced to $2^r + 2^{n-r} - 2$ (Keeney & Raiffa 1976 [10], Chapter 6.10.3). Finally, we point out that since the motivation of our work is to reduce the elicitation time, which in most cases is much larger than the computation time, this approach to reasoning with partial preference information could sometimes provide an attractive option despite the involved computational complexity.

Second, instead of working with only the supplied preferences, we may want the system to take the initiative and ask the user to make comparisons between decision consequences, or sets of decision consequences (using some qualitative logical constructs). To this end, the most interesting issue is identifying the questions whose answers would lead to the most conclusive inference, e.g., induced optimality of a particular decision alternative, or induced sub-optimality of a large number of decision alternatives [5]. A difficulty with this approach is that the user may not be able to answer some of those queries.

Finally, it is interesting to see if this approach to reasoning with partially elicited preference information can be intergrated into the case-based framework to preference elicitation advocated in our recent work [8] (see also [4]). In this framework, the system maintains a population of users with their preferences partially or completely specified in a given domain. When encountering a new user, called $A$, the system first elicits some preference information from $A$, and then determines which user in the population has the preference structure that is closest to $A$'s. The preference structure of that user will be used to determine an initial default representation (or working model) of $A$'s preferences. We can use the techniques presented in this paper to elicit some initial preference information from the decision maker and to eliminate some sub-optimal decision alternatives. If the set of remaining decision candidates is still large, the system can use the case-based approach to make recommendations for the decision maker.

---

[5] Our previous work [7] has addressed this issue in a special case when the utility function is additive.

### Acknowledgements

This work was partially supported by a UWM Graduate School Fellowship, by NSF grant IRI-9509165, and by United States Air Force, contact No. F30602-98-1-0045. We thank Jeff Erickson for pointers to relevant literature in computational geometry.

### References

[1] F. Bacchus and A. Grove. Utility independence in a qualitative decision theory. In *Proceedings of the Fifth International Conference on Knowledge Representation and Reasoning (KR '96)*, 1996. Morgan Kaufmann.

[2] F. Bacchus and A. Grove. Independence and qualitative decision theory. In *Working Notes of the Stanford Spring Symposium on Qualitative Decision Theory*, Stanford, CA, 1997.

[3] C. Boutilier. Toward a logic for qualitative decision theory. In *Principles of Knowledge Representation and Reasoning: Proceedings of the Fourth International Conference (KR94)*. Morgan Kaufmann, San Mateo, CA, 1994.

[4] U. Chajewska, L. Getoor, J. Norman, and Y. Shahar. Utility elicitation as a classification problem. In *Proceedings of the Fourteenth Conference on Uncertainty in Artificial Intelligence*, July 1998. To appear.

[5] J. Doyle, Y. Shoham, and M. Wellman. A logic of relative desires (preliminary report). In *Proceedings of the 6th International Symposium on Methodologies for Intelligent Systems*, pages 16–31, 1991.

[6] J. Doyle and M. Wellman. Representing preferences as ceteris paribus comparatives. In S. Hanks, S. Russel, and M. Wellman, editors, *Working Notes of the AAAI Spring Symposium on Decision-Theoretic Planning*. 'Stanford, CA, 1994.

[7] V. Ha and P. Haddawy. Problem-focused incremental elicitation of multi-attribute utility models. In *Proceedings of the Thirteenth Conference on Uncertainty in Artificial Intelligence*, pages 215–222, August 1997.

[8] V. Ha and P. Haddawy. Towards case-based preference elicitation: Similarity measures on preference structures. In *Proceedings of the Fourteenth Conference on Uncertainty in Artificial Intelligence*, pages 193–201, July 1998.

[9] G. Hazen. Partial information, dominance, and potential optimality in multiattribute utility theory. *Operations research*, 34(2):296–310, 1986.

[10] R.L. Keeney and H. Raiffa. *Decisions with Multiple Objectives: Preferences and Value Tradeoffs*. Wiley, New York, 1976.

[11] M. Koksalan and P. Sagala. Interactive aproaches for discrete alternative multiple criteria decision making with monotone utility functions. *Management Science*, 41(7):1158–1171, 1995.

[12] P. Korhonen, J Wallenius, and S. Zionts. Solving the discrete multiple criteria problem using convex cones. *Management Science*, 30(11):1336–1345, 1984.

[13] S. Prasad, M. Karwan, and S. Zionts. Use of convex cones in interactive multiple objective decision making. *Management Science*, 43(5):723–734, 1997.

[14] R. Seidel. Convex hull computations. In Jacob E. Goodman and Joseph O'Rourke, editors, *Handbook of Discrete and Computational Geometry*, chapter 19, pages 361–376. CRC Press LLC, 1997.

[15] S. Tan and J. Pearl. Qualitative decision theory. In *Proceedings of the Twelfth National Conference on Artificial Intelligence*, pages 928–932, 1994.

[16] S. Zionts and J. Wallenius. An interactive programming method for solving the multiple criteria proble. *Management Science*, 22(6):652–633, 1976.